\PassOptionsToPackage{table}{xcolor}   
\documentclass{article}

\usepackage[sglblindworkshop, final]{neurips_2026}
\workshoptitle{AXIOM}   

\usepackage[utf8]{inputenc}
\usepackage[T1]{fontenc}
\usepackage{hyperref}
\usepackage{url}
\usepackage{booktabs}
\usepackage{amsfonts}
\usepackage{nicefrac}
\usepackage{microtype}
\usepackage{xcolor}
\usepackage{graphicx}
\usepackage{multirow}
\usepackage{amsmath}
\usepackage{array}

\newcommand{\JMag}{J_{\mathrm{Mag}}}
\newcommand{\JLogit}{J_{\mathrm{Logit}}}
\newcommand{\JBIMag}{\text{J-BI}_{\mathrm{Mag}}}
\newcommand{\JBILogit}{\text{J-BI}_{\mathrm{Logit}}}

\title{Output-Aware Block Influence: Jacobian-Lens Weighting for Depth Pruning}

\newcommand{\authorcell}[3]{\makebox[0.48\textwidth]{\begin{tabular}[t]{@{}c@{}}\textbf{#1}\\ #2\\ \texttt{#3}\end{tabular}}}
\author{%
  \normalfont
  \begin{tabular}{@{}c@{}c@{}}
    \noalign{\vskip 12pt}
    \authorcell{Kyle Lemoi}{Independent}{kplemoi@gmail.com} &
    \authorcell{Jonathan Yong}{Independent}{jonathanjjyong@gmail.com} \\ \noalign{\vskip 1.2em}
    \authorcell{Shubhankar Tripathy}{University of Massachusetts Amherst}{stripathy@umass.edu} &
    \authorcell{Priestley Fernandes}{University of the Cumberlands}{pfernandes62171@ucumberlands.edu} \\ \noalign{\vskip 1.2em}
    \authorcell{Anirban Majumder\thanks{Work done independently and unrelated to his role at Amazon.}}{Amazon Science}{anirbanmj16@gmail.com} &
    \authorcell{Zarreen Reza}{Algoverse Research}{zarreen@algoverseairesearch.org}
  \end{tabular}%
}

\begin{document}

\maketitle

\begin{abstract}


Depth pruning removes entire transformer blocks and therefore depends on a reliable criterion for identifying which blocks can be removed. Block Influence (BI) ranks blocks by how minimally they rotate the hidden state. However, local redundancy does not necessarily imply downstream irrelevance. We address this limitation using the Jacobian Lens (J-Lens) neural artifact, a map transporting perturbations at a layer to the final residual stream.  At a common sparsity level, pruning scores applying J-Lens transportation to $BI$, what we call $J-BI$, were found to improve post-pruning accuracy by up to 19 points over $BI$ while reducing damage to perplexity for Qwen3-8B, Llama-3.1-8B-IT, and Gemma-3-12B-IT. We also perform a single-block ablation study and use J-Lens probing to explain why the proposed scores improved performance.


\end{abstract}

\section{Introduction}
\label{sec:intro}

Modern large language models require significant computational resources for inference and deployment. Pruning reduces resource requirements by removing model components ranging from weights to blocks.  Structural pruning targeting blocks yields latency and memory use improvements without requiring special kernels for deployment.\citep{men2024shortgpt,gromov2024unreasonable}

ShortGPT \citep{men2024shortgpt} is a popular block removal approach.  It decides which block to remove using a score they call Block Influence (BI) measuring how far each block rotates the hidden state across a calibration dataset. This similarity score is local and does not measure downstream influence. A block might make a small direction change locally that later layers amplify or make a large local change which does not reach the output. BI does not account for this.

The Jacobian lens (J-Lens) \citep{gurnee2026workspace} is a linear map, fit once per model and available pretrained for many open models, that expresses how a change at any layer is expected to affect the final residual stream.  Other studies have used interpretability approaches to assess where pruner targeting might be useful (Appendix~\ref{related_work_section}). Here we directly incorporate the J-Lens' map to create scores for pruning.  

\begin{figure}[t]
    \centering
    \includegraphics[width=0.87\linewidth]{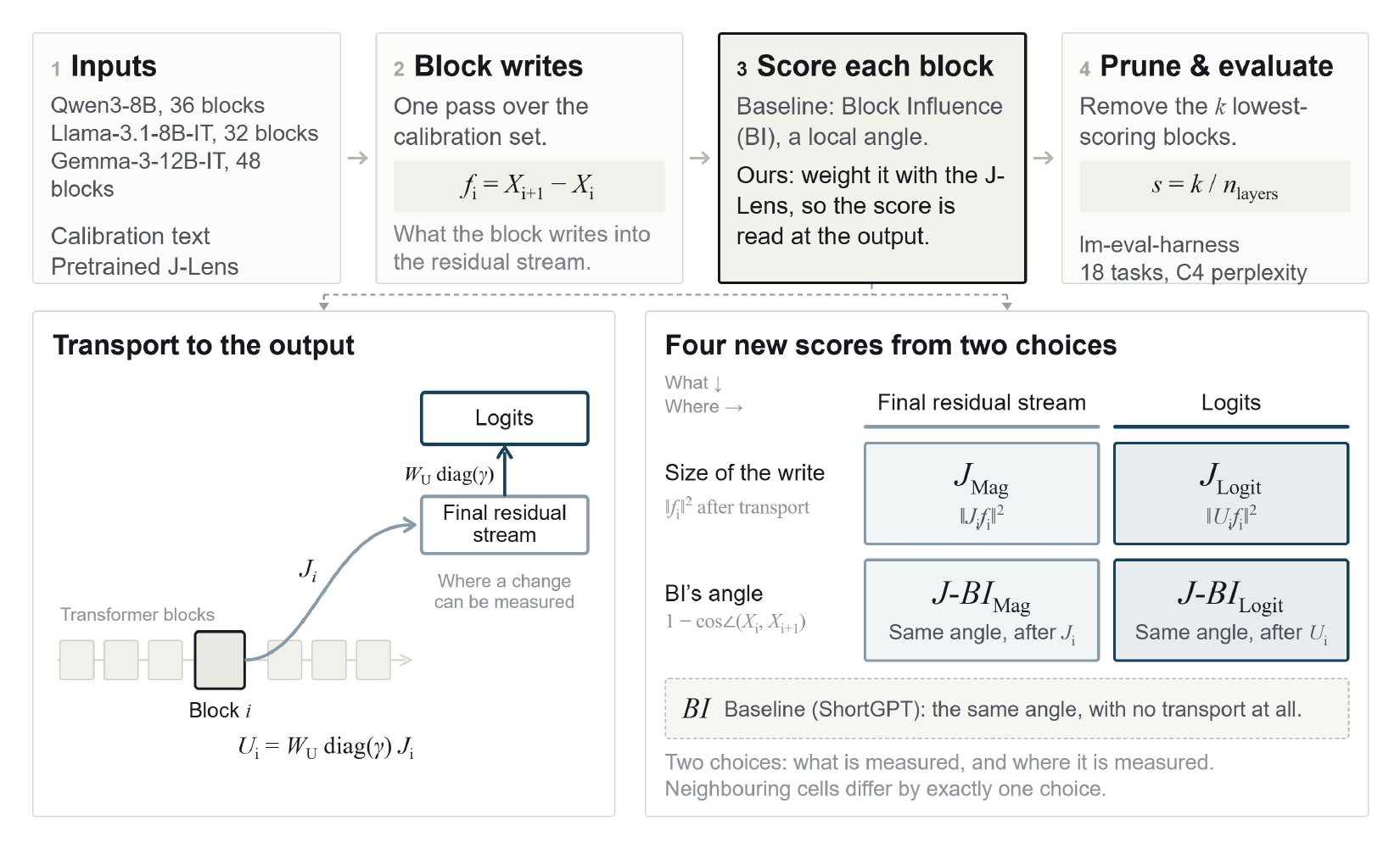}

    \caption{\textbf{Method overview.} \emph{Top:} Residual streams are recorded across a calibration set.  The streams are used to score each block. For each score the $k$ lowest-scoring blocks are removed and the pruned model evaluated. \emph{Bottom left:} the J-Lens $J_i$ carries a change at block $i$ to the final residual stream. The perturbation is carried further to the logits using $U_i = W_U\,\mathrm{diag}(\gamma)\,J_i$. \emph{Bottom right:} our four scores. Columns are where the change is measured, either at the final residual stream or the logits.  Rows are what is measured after transport, either the size or the angle change. BI is the untransported baseline.}
    
    \label{method_overview}
\end{figure}

ShortGPT's similarity-based measure considers only the local similarity between a block's inputs and outputs. The J-Lens provides a way to assess the impact of a local perturbation on the final residual stream and logits.  We use this fact to propose two classes of scores.  The first directly uses the J-Lens' projected impact on the final residual stream and logits to select blocks for pruning. The second class of scores, $J-BI$, combines the local layer similarity of $BI$ with the J-Lens transport to create similarity scores at final residual stream and logits. Figure \ref{method_overview} shows the scores and where they sit in the pruning pipeline. 

Our findings are twofold. First, we propose a way of creating pruning scores based upon a recently released mechanical interpretation tool, the J-Lens. We find that direct use of the J-lens is not a sufficient signal for pruning.  However applying the J-Lens' transport to BI similarity improves both accuracy and perplexity.  Second, we provide an explanation for the qualitatively different behavior observed across the pruning scores.  We do this through two assessments, a study of perplexity changes under single block ablation and through use of the J-Lens as originally intended, an interpretation tool for studying the thinking space of models. 

\section{Method}
\label{sec:method}

\paragraph{Block Influence.}
ShortGPT \citep{men2024shortgpt} removes whole transformer blocks. Each block $i$ receives a score, Block Influence (BI), from a single pass over a calibration set:
\begin{equation}
  \mathrm{BI}_i \;=\; 1 - \mathbb{E}_{X,t}\!\left[
    \frac{X_{i,t}^{\top} X_{i+1,t}}
         {\lVert X_{i,t}\rVert_2\,\lVert X_{i+1,t}\rVert_2}
  \right],
  \label{eq:bi}
\end{equation}
where $X_{i,t}$ is the hidden state entering block $i$ at token $t$ and $X_{i+1,t}$ is the state leaving it. BI is based upon cosine similarity between a block's input and output. A block that computes a relatively small rotation of the hidden state gets a low score, and the $k$ lowest-scoring blocks are removed.

\paragraph{The Jacobian lens.}
The J-Lens \citep{gurnee2026workspace} is a linear map from the residual stream at layer $\ell$ to the final residual stream, $J_\ell = \mathbb{E}_{t,\,t'\ge t,\,\mathrm{prompt}}\big[\partial h_{\mathrm{final},t'} / \partial h_{\ell,t}\big]$: the average, over a text corpus, of the first-order effect a small change at layer $\ell$ has on the final layer. Multiplying a change at layer $\ell$ by $J_\ell$ gives its expected effect on the final residual stream; multiplying by $U_\ell = W_U\,\mathrm{diag}(\gamma)\,J_\ell$, where $\gamma$ is the final RMSNorm gain and $W_U$ the unembedding, carries it on to the logits.

\paragraph{Four proposed scores}
Removing block $i$ deletes its contribution to the residual stream $f_{i,t} = X_{i+1,t} - X_{i,t}$, at every token. BI computes the hidden state rotation. J-BI first transports the block contribution to the output layer (or logits) and then computes hidden state rotation at the output layer (or at the logits).

We define $T_i = J_i$ for transport to the final residual stream and $T_i = U_i$ for transport to the logits. Each defines a metric $M_i = T_i^{\top}T_i$, and $\delta^{\top} M_i \delta = \|T_i\delta\|^2$ is the squared size of a change $\delta$ after transport (Appendix~\ref{app:derivation}). The two \emph{lens-only} scores, $\JMag$ and $\JLogit$, rank blocks by the expected transported size of their residual update, $\mathbb{E}_{X,t}\|T_i f_{i,t}\|^2$. They measure a block's impact at the output, but discard the local similarity that BI relies on.

\paragraph{J-BI: Block Influence under a lens metric.}
BI's cosine is built from the ordinary inner product $u^{\top}v$. Any metric $M$ gives a new inner product $u^{\top} M v$, and with it a new cosine. With $M = I$ this is BI. With $M = M_i$ it is $(T_i u)^{\top}(T_i v)$: the same angle, measured after transport. So
\begin{equation}
  \text{J-BI}_i \;=\; 1 - \mathbb{E}_{X,t}\!\left[
    \frac{(T_i X_{i,t})^{\top}(T_i X_{i+1,t})}
         {\|T_i X_{i,t}\|_2\,\|T_i X_{i+1,t}\|_2}\right],
  \label{eq:jbi}
\end{equation}
which gives $\JBIMag$ for $T_i = J_i$ and $\JBILogit$ for $T_i = U_i$. With BI as the reference, the five scores form the grid in Figure~\ref{method_overview}; all need only BI's calibration pass plus one matrix product per block.

\paragraph{Experimental setup.}
We evaluate Qwen3-8B (36 blocks), Llama-3.1-8B-IT (32) and Gemma-3-12B-IT (48), using the same calibration text and the pretrained lens of \citet{gurnee2026workspace} for every score. We remove the $k$ lowest-scoring blocks for $k=1,2,4,\dots,28$, report depth sparsity $s = k/n_{\mathrm{layers}}$, and evaluate 18 zero-shot tasks (\texttt{lm-eval-harness}) \citep{phang2022eleutheraigoingopenscience} and C4 perplexity \citep{jelinek1977perplexity}. Accuracy is a filtered macro-average, where a task counts toward a model's average only if the dense model scores at least 5 points on it, which drops tasks already at floor (details and per-task values in Appendices~\ref{experiments_section} and~\ref{app:pertask}).

\section{Results}
\label{sec:results}

\begin{figure}[t]
    \centering
    \includegraphics[width=\linewidth]{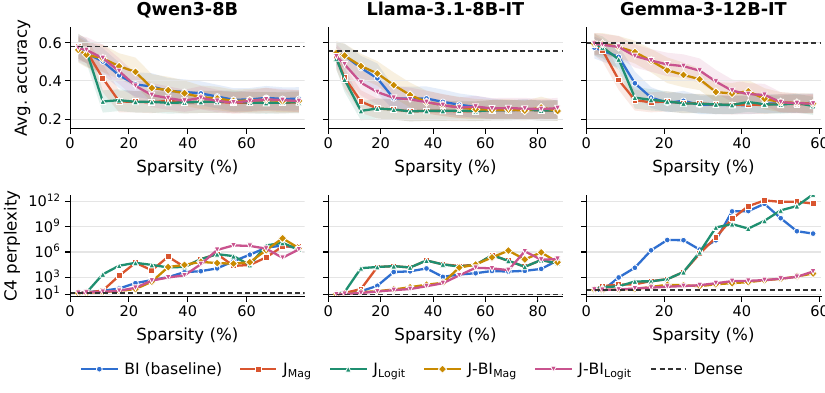}
    \caption{Average accuracy over 18 tasks (\textbf{top}; bands are $\pm 1$ standard error across tasks) and C4 perplexity (\textbf{bottom}; log scale, lower is better) against depth sparsity $s = k/n_\mathrm{layers}$, $k = 1,\dots,28$. Above roughly $10^{3}$ the model has collapsed. Dense (the dashed line) is the unpruned model performance.}
    \label{stacked_acc_ppl}
\end{figure}

\begin{table}[t]
  \caption{Results with $k=8$ blocks removed ($s = 22\%$, $25\%$, $17\%$). Accuracy is the filtered macro-average over 18 tasks (\%); C4 perplexity is lower-is-better. \emph{Dense} (the unpruned model) and BI are the reference points; the four lens scores are ours. \textbf{Bold}: best pruning score per row. $\JMag$ and $\JLogit$ select the same blocks on Qwen and Llama.}
  \label{tab:main-k8}
  \centering
  \footnotesize
  \setlength{\tabcolsep}{5pt}
  \begin{tabular}{@{}l l r r @{\hspace{1.6em}} r r r r@{}}
    \toprule
    & & \multicolumn{2}{c@{\hspace{1.6em}}}{Reference} & \multicolumn{2}{c}{Lens only} & \multicolumn{2}{c@{}}{J-BI} \\
    \cmidrule(lr){3-4}\cmidrule(lr){5-6}\cmidrule(l){7-8}
    Model & & Dense & BI & $\JMag$ & $\JLogit$ & $\JBIMag$ & $\JBILogit$ \\
    \midrule
    Qwen3-8B        & Acc.\ $\uparrow$ & 58.1 & 38.2 & 29.1 & 29.1 & \textbf{44.7} & 37.4 \\
                    & Ppl.\ $\downarrow$ & 13.2 & 202 & 52,330 & 52,330 & \textbf{40.7} & 57.2 \\
    \addlinespace[1pt]
    Llama-3.1-8B-IT & Acc.\ $\uparrow$ & 55.7 & 30.9 & 25.0 & 25.0 & \textbf{37.9} & 31.2 \\
                    & Ppl.\ $\downarrow$ & 8.8 & 4,204 & 21,163 & 21,163 & 35.8 & \textbf{30.0} \\
    \addlinespace[1pt]
    Gemma-3-12B-IT  & Acc.\ $\uparrow$ & 60.0 & 31.3 & 28.8 & 30.1 & 50.6 & \textbf{50.7} \\
                    & Ppl.\ $\downarrow$ & 33.4 & $1.58\!\times\!10^{6}$ & 354 & 333 & 72.5 & \textbf{60.3} \\
    \bottomrule
  \end{tabular}
\end{table}

\paragraph{Measuring BI at the output improves pruning.}
Across all three models, $\JBIMag$ beats BI in accuracy when 4–8 blocks are removed, while maintaining lower perplexity until both methods collapse (Figure~\ref{stacked_acc_ppl}). At eight blocks (Table~\ref{tab:main-k8}) it gains approximately 7 points on both Qwen and Llama, and 19 on Gemma while lowering C4 perplexity by $5\times$, $117\times$ and $26,000\times$ on Qwen, Llama, and Gemma, respectively. On Gemma this is a change of regime, where BI's perplexity passes $10^{3}$ by six blocks removed (12\% sparsity); $\JBIMag$'s stays below $10^{3}$ through 24 blocks (50\%). Per task (Table~\ref{tab:acc_k8}), $\JBIMag$ is the best score on most tasks for every model.

\begin{figure}[t]
  \centering
  \includegraphics[width=1.0\linewidth]{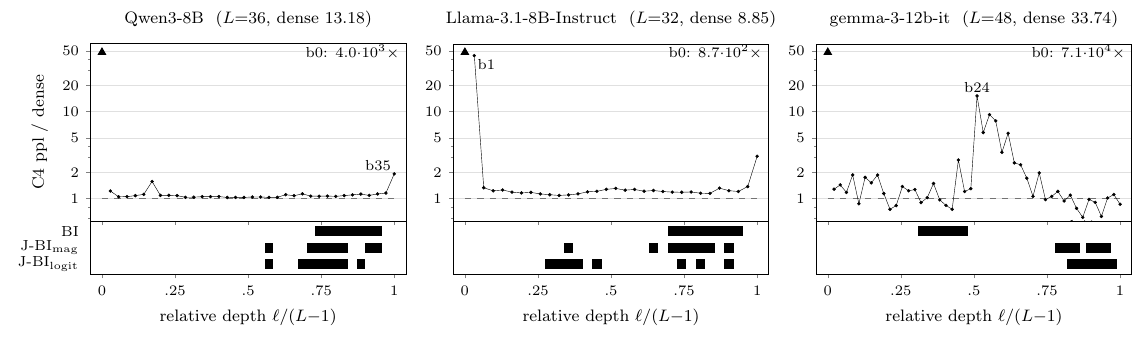}
  \caption{\textbf{Single-block sensitivity and selected removals.} \emph{Top:} C4 perplexity after removing one block, relative to dense (dotted line); off-scale values annotated. \emph{Bottom:} blocks removed at $k=8$ by BI and the two J-BI scores. BI removes Gemma's sensitive middle blocks; J-BI does not.}
  \label{fig:single-block}
\end{figure}

\paragraph{Lens magnitude is not a sufficient pruning signal}
$\JMag$ and $\JLogit$ fall below BI in accuracy on all three models (Table~\ref{tab:main-k8}) and collapse earlier, by four to six blocks on Qwen and Llama (Figure~\ref{stacked_acc_ppl}). At $k=8$ they also prune shallow blocks (mean depth 17\%), where removing a single block can result in perplexity growing by greater than $10^2$ (Figure~\ref{fig:single-block}). Therefore, the perplexity improvement comes from keeping BI's local similarity and measuring the angle where downstream sensitivity is visible. Of the two J-BI scores, $\JBIMag$ is the more reliable on accuracy ($\JBILogit$ ties $\JBIMag$ on Gemma but falls below BI on Llama at $k=2$ to $6$). On perplexity, $\JBILogit$ is close to $\JBIMag$, and $\JBILogit$ is lower than $\JBIMag$ on Llama and Gemma at $k=8$.

\paragraph{Gemma shows the failure mode of local BI.}
Removing one block at a time (Figure~\ref{fig:single-block}, top) locates a group of middle Gemma blocks, roughly 20 to 28, whose removal increases perplexity by up to a factor of 20. BI removes from this region at $k=8$ because the states entering and leaving these blocks are cosine-similar; J-BI leaves them alone and prunes later blocks (Figure~\ref{fig:single-block}, bottom). For Qwen and Llama, BI and $\JBIMag$ agree on three quarters of the removed blocks, and the improvements in perplexity are relatively smaller. A lens readout (Appendix~\ref{jbi_arch_agnostic}) shows that these middle Gemma blocks (20 to 28) are where Gemma's next-token predictions first become readable, illustrating that a block can appear redundant locally while remaining consequential at the output. We hypothesize that Gemma's alternating local and global attention may contribute to the pruning sensitivity to perplexity in this middle region.

\section{Conclusions and Limitations}

$J-BI$ makes a local pruning score output-aware by transporting the local Block Influence ($BI$) pruning signal to the final residual stream. On three model families $J-BI$ improved both accuracy and perplexity over $BI$. Scores based only on transport magnitude, $J_{Mag}$ and $J_{Logit}$, maintained or degraded performance. These results show that $BI$'s transported similarity was responsible for the improvement rather than the magnitude of the lens. We used an ablation study and J-Lens probing to explain the realized gains from a mechanistic interpretability perspective.  The current work is limited in that it did not assess performance across model scales or assess realized changes in throughput.  Also, our study relied upon pretrained J-Lens. Costs of training a J-Lens from scratch were not considered. In future work we intend to fill the scale gap as well as extend the J-Lens based score weighting to other pruning algorithms and paradigms including kv-cache pruning.

\bibliographystyle{plainnat}
\bibliography{references}

\appendix


\section{Related Work}
\label{related_work_section}

\paragraph{Pruning and the importance score.}
Pruning shrinks a network by removing part of it. The central design choice in any pruning method is the score that decides what to remove. Early methods removed individual weights with the smallest magnitude \citep{han2015learning}.  However magnitude alone did not hold up as models reached the current LLM scale. SparseGPT \citep{frantar2023sparsegpt} uses second-order information and updates the remaining weights to compensate for each removal. Wanda \citep{sun2024wanda} scores each weight by its magnitude times the norm of its input activation. Later work adds gradient information to better estimate the damage a removal causes, e.g.\ Wanda++ \citep{yang2025wandapp}. All of these produce unstructured or semi-structured sparsity, which is hard to turn into wall-clock gains on standard hardware. Structured methods remove whole units instead: coupled heads and neurons \citep{ma2023llmpruner}, rows and columns after a PCA rotation \citep{ashkboos2024slicegpt}, or units chosen jointly with retraining or distillation
\citep{xia2024sheared,sreenivas2024minitron}.

\paragraph{Depth pruning.}
The coarsest structured approach removes entire transformer blocks, which gives a proportional speedup with no special kernels. ShortGPT \citep{men2024shortgpt} scores each block by
Block Influence (BI): one minus the cosine similarity between the hidden states entering and leaving the block, averaged over a calibration set. BI is calculated and the lowest-scoring blocks are removed. \citet{gromov2024unreasonable} use a similar angular distance between layers $\ell$ and $\ell+n$, then fine-tune briefly. SLEB \citep{song2024sleb} and Shortened LLaMA \citep{kim2024shortened} instead measure the damage directly. They remove each block in turn and record the change in perplexity or loss.  However, this costs a forward sweep per block. LaCo \citep{yang2024laco} merges neighbouring layers rather than deleting them. Across these approaches the scores are either local - assessing how much hidden states change in a layer neighborhood - or interventional, measuring how much some output measure suffers when a block is removed.  There is an unoccupied middle-ground: given how a block changes the hidden state, how much of that change reaches the output?  This space is where our proposed scores live.

\paragraph{Reading internal states with lenses.}
Mechanistic interpretability has built tools that read off what an internal activation will do to the output. The logit lens \citep{nostalgebraist2020logitlens} applies the unembedding directly to an intermediate residual stream and the tuned lens \citep{belrose2023tunedlens} fits a small affine translator per layer to make that readout more
faithful. The Jacobian lens (J-Lens) \citep{gurnee2026workspace} goes further. It fits the average Jacobian of the final residual stream with respect to each layer's residual stream over a text corpus.  This process gives a fixed linear map that transports a change at layer $\ell$ to its expected effect at the output. \citet{gurnee2026workspace} use the lens to find a ``workspace'' of verbalizable middle layer representations.  Interpretability work has also shown that particular components, such as persona vectors \citep{chen2025persona} and task circuits \citep{conmy2023acdc}, carry causal weight for behaviour and that ablating them changes what the model does. A pruning score should therefore know which directions of change matter to the output. To this end we take the J-Lens map and use it as a ruler with which to measure downstream contributions of a block.

\paragraph{Interpretability-guided pruning.}
A small line of recent work uses interpretability signals to decide what to prune. \citet{hatefi2025attribution} use layer-wise relevance propagation to score neurons and weights, for compression and for targeted removal of behaviours. AAPP \citep{patel2025aapp} reserves alignment-critical channels during dynamic pruning, and
Persona-Pruner \citep{kim2026personapruner} learns feed-forward masks that keep a persona-specific sub-network. All of these approaches score by attribution, ablation, or learned masks. None of them use a lens as the criterion and none target whole-block removal. Our work keeps ShortGPT's procedure unchanged and changes only the geometry inside its score.  We use the J-Lens to transport a block's input and output to the final residual stream, or on to the logits, before similarities are measured. Blocks are then judged by whether their changes survive to the output rather than by changes to their local activations. To our knowledge this is the first use of a lens as a pruning criterion.
\section{Experimental details}
\label{experiments_section}

\paragraph{Models.}
We evaluate Qwen3-8B (36 blocks), Llama-3.1-8B-IT (32 blocks) and Gemma-3-12B-IT (48 blocks). They give a spread across three model families at a comparable scale, and each is a dense model with a pretrained J-Lens available \citep{gurnee2026workspace}. Newer models with pretrained lenses exist, but they use mixture-of-experts architectures, which would confound the interpretation of block removal. For this workshop we rely on dense models to establish the proof of concept.

\paragraph{Scoring.}
Every score uses the same calibration pass as BI. For block $i$ we use the lens fit at the block's output layer. Expectations over tokens exclude the first 16 positions of each sequence and the final position. For the J-BI scores each state is transported once and an ordinary cosine is taken; the score is the mean of per-token cosines. The lens-only scores are the mean over tokens of the transported residual update energy $f_{i,t}^{\top} M f_{i,t}$.

\paragraph{Tasks.}
Following prior depth-pruning work \citep{men2024shortgpt,kim2024shortened}, we assess pruned models on a mixture of zero-shot task accuracy and perplexity. The 18 tasks cover several capabilities:
\begin{description}
  \item[Commonsense and language reasoning.] HellaSwag, PIQA, WinoGrande, BoolQ, RTE.
  \item[Multi-step reasoning.] GSM8K (with and without chain-of-thought), DROP, StrategyQA.
  \item[Multi-hop question answering.] HotpotQA, MuSiQue.
  \item[Knowledge and science QA.] MMLU, TriviaQA, ARC-Easy, ARC-Challenge, OpenBookQA.
  \item[Bias and truthfulness.] BBQ, TruthfulQA.
  \item[Language modelling.] Perplexity on C4, measured over 2048-token windows.
\end{description}

\paragraph{Assessment.}
We compare BI with $\JMag$, $\JLogit$, $\JBIMag$ and $\JBILogit$ on every model and task using the EleutherAI \texttt{lm-eval-harness}. The number of removed blocks $k$ is swept over $1, 2, 4, \dots, 28$ and converted to depth sparsity $s = k/n_{\mathrm{layers}}$ for plotting. The filtered macro-average admits a task into a model's average only when the dense model scores at least 5 points on it; this keeps tasks at floor (for example MuSiQue) from adding noise. The bands in Figure~\ref{stacked_acc_ppl} are $\pm 1$ standard error of the mean across admitted tasks.

\section{Metric derivation}
\label{app:derivation}

Removing block $i$ injects the perturbation $-f_{i,t}$, where $f_{i,t} = X_{i+1,t} - X_{i,t}$ is the block's residual update, into the residual stream at the block's position. To first order, the lens carries a change $\delta$ at that position to a displacement $J_i\delta$ in the final residual stream and $U_i\delta = W_U\,\mathrm{diag}(\gamma)\,J_i\delta$ in the logits. The squared size of each displacement is a quadratic form in $\delta$:


\begin{align}
  \|J_i\delta\|^2 &= \delta^{\top} J_i^{\top} J_i\,\delta = \delta^{\top} \JMag\,\delta, \label{eq:quadform}\\
  \|U_i\delta\|^2 &= \delta^{\top} J_i^{\top}\,\mathrm{diag}(\gamma)\,W_U^{\top}W_U\,\mathrm{diag}(\gamma)\,J_i\,\delta = \delta^{\top} \JLogit\,\delta . \label{eq:quadform-logit}
\end{align}
Both metrics are positive semidefinite. $\JMag$ is the Gram matrix of the lens: its eigenvectors are the right singular directions of $J_i$ and its eigenvalues the squared singular values, so a component of $\delta$ that the network transmits strongly costs more and a component in the lens's null space costs nothing. $\JLogit$ pushes the same measurement through the final RMSNorm gain and the unembedding. It omits the RMSNorm's $1/\|h\|$ scale, which cancels within a layer's ranking, and its radial projector, which would need a forward pass.

The lens-only scores average the quadratic form over the calibration set, $\mathbb{E}_{X,t}[\,f_{i,t}^{\top} M f_{i,t}\,]$, and so estimate the first-order damage of deleting the block. J-BI uses the same $M$ inside BI's cosine instead: $(J_i X_{i,t})^{\top}(J_i X_{i+1,t}) = X_{i,t}^{\top}\JMag X_{i+1,t}$ and likewise for $U_i$ and $\JLogit$, so Equation~\ref{eq:jbi} is exactly Equation~\ref{eq:bi} with the inner product $u^{\top}v$ replaced by $u^{\top} M v$. Setting $M = I$ recovers BI.

\section{Per-task accuracy at $k=8$}
\label{app:pertask}

Table~\ref{tab:acc_k8} lists every task for every model and score with eight blocks removed. The filtered macro-average in the last column is the number reported in Table~\ref{tab:main-k8}.

%
%
%
%
%
\makeatletter
\@ifundefined{cellcolor}{%
  \PackageWarningNoLine{k8-table}{Row shading is OFF: \string\cellcolor\space
    is undefined, so xcolor's `table' option (colortbl) is not loaded.
    Fix with \string\PassOptionsToPackage{table}{xcolor} before
    \string\documentclass, or \string\usepackage{colortbl} after xcolor}%
  \providecommand{\cellcolor}[1]{}%
}{}
\@ifundefined{definecolor}{%
  \PackageWarningNoLine{k8-table}{Colour is OFF: xcolor is not loaded}%
  \providecommand{\definecolor}[3]{}%
  \providecommand{\textcolor}[2]{#2}%
}{}
\makeatother
\definecolor{refgray}{gray}{0.45}
\definecolor{ourstint}{HTML}{EEF3FA}
\begin{table}[t]
\centering
\renewcommand{\arraystretch}{1.3}
\resizebox{\textwidth}{!}{%
\begin{tabular}{l@{\hspace{5pt}}c@{\hspace{6pt}}lrrrrrrrrrrrrrrrrrr@{\hspace{9pt}}r}
\toprule
Model &  & Metric & \rotatebox[origin=l]{90}{ARC-c} & \rotatebox[origin=l]{90}{ARC-e} & \rotatebox[origin=l]{90}{BBQ} & \rotatebox[origin=l]{90}{BoolQ} & \rotatebox[origin=l]{90}{DROP} & \rotatebox[origin=l]{90}{GSM8K (CoT)} & \rotatebox[origin=l]{90}{GSM8K (no-CoT)} & \rotatebox[origin=l]{90}{HellaSwag} & \rotatebox[origin=l]{90}{HotpotQA} & \rotatebox[origin=l]{90}{MMLU} & \rotatebox[origin=l]{90}{MuSiQue} & \rotatebox[origin=l]{90}{OpenBookQA} & \rotatebox[origin=l]{90}{PIQA} & \rotatebox[origin=l]{90}{RTE} & \rotatebox[origin=l]{90}{StrategyQA} & \rotatebox[origin=l]{90}{TriviaQA} & \rotatebox[origin=l]{90}{TruthfulQA} & \rotatebox[origin=l]{90}{WinoGrande} & \rotatebox[origin=l]{90}{\textbf{Average}} \\
\midrule
\addlinespace[1.5pt]
\multirow{6}{*}{\begin{tabular}[c]{@{}l@{}}Qwen3-8B\\ $s = 22.2\%$\end{tabular}} & \multirow{2}{*}{\rotatebox[origin=c]{90}{\scriptsize Baseline}} & \textcolor{refgray}{\textit{Dense}} & 0.54 & 0.83 & 0.46 & 0.86 & 0.39 & 0.94 & 0.18 & 0.55 & 0.05 & 0.71 & 0.00 & 0.31 & 0.77 & 0.78 & 0.68 & 0.47 & 0.27 & 0.69 & 0.58 \\
 &  & BI & 0.28 & 0.45 & \textbf{0.39} & 0.66 & 0.11 & 0.00 & 0.02 & \textbf{0.37} & 0.01 & 0.42 & 0.00 & 0.20 & 0.60 & 0.68 & 0.52 & 0.02 & 0.24 & 0.54 & 0.38 \\
\addlinespace[1.5pt]
\cmidrule(lr){2-22}
\addlinespace[1.5pt]
 & \cellcolor{ourstint}{} & \cellcolor{ourstint}{J\textsubscript{Mag}} & \cellcolor{ourstint}{0.21} & \cellcolor{ourstint}{0.28} & \cellcolor{ourstint}{\textbf{0.39}} & \cellcolor{ourstint}{0.38} & \cellcolor{ourstint}{0.00} & \cellcolor{ourstint}{0.00} & \cellcolor{ourstint}{0.00} & \cellcolor{ourstint}{0.24} & \cellcolor{ourstint}{0.00} & \cellcolor{ourstint}{0.27} & \cellcolor{ourstint}{0.00} & \cellcolor{ourstint}{0.17} & \cellcolor{ourstint}{0.53} & \cellcolor{ourstint}{0.52} & \cellcolor{ourstint}{0.53} & \cellcolor{ourstint}{0.00} & \cellcolor{ourstint}{0.22} & \cellcolor{ourstint}{0.48} & \cellcolor{ourstint}{0.29} \\
 & \cellcolor{ourstint}{} & \cellcolor{ourstint}{J\textsubscript{Logit}} & \cellcolor{ourstint}{0.21} & \cellcolor{ourstint}{0.28} & \cellcolor{ourstint}{\textbf{0.39}} & \cellcolor{ourstint}{0.38} & \cellcolor{ourstint}{0.00} & \cellcolor{ourstint}{0.00} & \cellcolor{ourstint}{0.00} & \cellcolor{ourstint}{0.24} & \cellcolor{ourstint}{0.00} & \cellcolor{ourstint}{0.27} & \cellcolor{ourstint}{0.00} & \cellcolor{ourstint}{0.17} & \cellcolor{ourstint}{0.53} & \cellcolor{ourstint}{0.52} & \cellcolor{ourstint}{0.53} & \cellcolor{ourstint}{0.00} & \cellcolor{ourstint}{0.22} & \cellcolor{ourstint}{0.48} & \cellcolor{ourstint}{0.29} \\
 & \cellcolor{ourstint}{} & \cellcolor{ourstint}{J-BI\textsubscript{Mag}} & \cellcolor{ourstint}{\textbf{0.33}} & \cellcolor{ourstint}{\textbf{0.57}} & \cellcolor{ourstint}{\textbf{0.39}} & \cellcolor{ourstint}{\textbf{0.83}} & \cellcolor{ourstint}{0.11} & \cellcolor{ourstint}{\textbf{0.03}} & \cellcolor{ourstint}{\textbf{0.04}} & \cellcolor{ourstint}{\textbf{0.37}} & \cellcolor{ourstint}{\textbf{0.03}} & \cellcolor{ourstint}{\textbf{0.67}} & \cellcolor{ourstint}{0.00} & \cellcolor{ourstint}{\textbf{0.24}} & \cellcolor{ourstint}{\textbf{0.62}} & \cellcolor{ourstint}{0.71} & \cellcolor{ourstint}{\textbf{0.60}} & \cellcolor{ourstint}{\textbf{0.04}} & \cellcolor{ourstint}{\textbf{0.27}} & \cellcolor{ourstint}{\textbf{0.64}} & \cellcolor{ourstint}{\textbf{0.45}} \\
 & \cellcolor{ourstint}{\multirow{-4}{*}{\rotatebox[origin=c]{90}{\scriptsize \bfseries Ours}}} & \cellcolor{ourstint}{J-BI\textsubscript{Logit}} & \cellcolor{ourstint}{0.30} & \cellcolor{ourstint}{0.45} & \cellcolor{ourstint}{0.34} & \cellcolor{ourstint}{0.53} & \cellcolor{ourstint}{\textbf{0.12}} & \cellcolor{ourstint}{0.01} & \cellcolor{ourstint}{0.03} & \cellcolor{ourstint}{0.36} & \cellcolor{ourstint}{0.02} & \cellcolor{ourstint}{0.31} & \cellcolor{ourstint}{0.00} & \cellcolor{ourstint}{0.22} & \cellcolor{ourstint}{0.61} & \cellcolor{ourstint}{\textbf{0.77}} & \cellcolor{ourstint}{0.56} & \cellcolor{ourstint}{0.01} & \cellcolor{ourstint}{0.24} & \cellcolor{ourstint}{0.57} & \cellcolor{ourstint}{0.37} \\
\addlinespace[1.5pt]
\midrule
\addlinespace[1.5pt]
\multirow{6}{*}{\begin{tabular}[c]{@{}l@{}}Llama-3.1-8B-IT\\ $s = 25.0\%$\end{tabular}} & \multirow{2}{*}{\rotatebox[origin=c]{90}{\scriptsize Baseline}} & \textcolor{refgray}{\textit{Dense}} & 0.49 & 0.82 & 0.48 & 0.84 & 0.22 & 0.67 & 0.10 & 0.58 & 0.07 & 0.66 & 0.00 & 0.35 & 0.81 & 0.74 & 0.68 & 0.64 & 0.29 & 0.75 & 0.56 \\
 &  & BI & 0.30 & 0.46 & 0.40 & 0.38 & 0.00 & 0.00 & 0.02 & 0.33 & 0.00 & 0.32 & 0.00 & 0.22 & 0.67 & \textbf{0.66} & 0.52 & 0.01 & 0.28 & 0.65 & 0.31 \\
\addlinespace[1.5pt]
\cmidrule(lr){2-22}
\addlinespace[1.5pt]
 & \cellcolor{ourstint}{} & \cellcolor{ourstint}{J\textsubscript{Mag}} & \cellcolor{ourstint}{0.20} & \cellcolor{ourstint}{0.26} & \cellcolor{ourstint}{0.39} & \cellcolor{ourstint}{0.46} & \cellcolor{ourstint}{0.00} & \cellcolor{ourstint}{0.00} & \cellcolor{ourstint}{0.00} & \cellcolor{ourstint}{0.23} & \cellcolor{ourstint}{0.00} & \cellcolor{ourstint}{0.25} & \cellcolor{ourstint}{0.00} & \cellcolor{ourstint}{0.14} & \cellcolor{ourstint}{0.54} & \cellcolor{ourstint}{0.48} & \cellcolor{ourstint}{0.53} & \cellcolor{ourstint}{0.00} & \cellcolor{ourstint}{0.23} & \cellcolor{ourstint}{0.52} & \cellcolor{ourstint}{0.25} \\
 & \cellcolor{ourstint}{} & \cellcolor{ourstint}{J\textsubscript{Logit}} & \cellcolor{ourstint}{0.20} & \cellcolor{ourstint}{0.26} & \cellcolor{ourstint}{0.39} & \cellcolor{ourstint}{0.46} & \cellcolor{ourstint}{0.00} & \cellcolor{ourstint}{0.00} & \cellcolor{ourstint}{0.00} & \cellcolor{ourstint}{0.23} & \cellcolor{ourstint}{0.00} & \cellcolor{ourstint}{0.25} & \cellcolor{ourstint}{0.00} & \cellcolor{ourstint}{0.14} & \cellcolor{ourstint}{0.54} & \cellcolor{ourstint}{0.48} & \cellcolor{ourstint}{0.53} & \cellcolor{ourstint}{0.00} & \cellcolor{ourstint}{0.23} & \cellcolor{ourstint}{0.52} & \cellcolor{ourstint}{0.25} \\
 & \cellcolor{ourstint}{} & \cellcolor{ourstint}{J-BI\textsubscript{Mag}} & \cellcolor{ourstint}{\textbf{0.35}} & \cellcolor{ourstint}{0.58} & \cellcolor{ourstint}{\textbf{0.44}} & \cellcolor{ourstint}{\textbf{0.68}} & \cellcolor{ourstint}{\textbf{0.06}} & \cellcolor{ourstint}{\textbf{0.05}} & \cellcolor{ourstint}{\textbf{0.04}} & \cellcolor{ourstint}{\textbf{0.44}} & \cellcolor{ourstint}{\textbf{0.02}} & \cellcolor{ourstint}{\textbf{0.52}} & \cellcolor{ourstint}{0.00} & \cellcolor{ourstint}{\textbf{0.24}} & \cellcolor{ourstint}{\textbf{0.73}} & \cellcolor{ourstint}{0.65} & \cellcolor{ourstint}{\textbf{0.56}} & \cellcolor{ourstint}{\textbf{0.04}} & \cellcolor{ourstint}{\textbf{0.29}} & \cellcolor{ourstint}{\textbf{0.66}} & \cellcolor{ourstint}{\textbf{0.38}} \\
 & \cellcolor{ourstint}{\multirow{-4}{*}{\rotatebox[origin=c]{90}{\scriptsize \bfseries Ours}}} & \cellcolor{ourstint}{J-BI\textsubscript{Logit}} & \cellcolor{ourstint}{0.28} & \cellcolor{ourstint}{\textbf{0.62}} & \cellcolor{ourstint}{0.30} & \cellcolor{ourstint}{0.46} & \cellcolor{ourstint}{0.00} & \cellcolor{ourstint}{0.01} & \cellcolor{ourstint}{0.02} & \cellcolor{ourstint}{0.43} & \cellcolor{ourstint}{0.00} & \cellcolor{ourstint}{0.26} & \cellcolor{ourstint}{0.00} & \cellcolor{ourstint}{0.23} & \cellcolor{ourstint}{0.71} & \cellcolor{ourstint}{0.53} & \cellcolor{ourstint}{0.55} & \cellcolor{ourstint}{0.03} & \cellcolor{ourstint}{0.21} & \cellcolor{ourstint}{0.58} & \cellcolor{ourstint}{0.31} \\
\addlinespace[1.5pt]
\midrule
\addlinespace[1.5pt]
\multirow{6}{*}{\begin{tabular}[c]{@{}l@{}}Gemma-3-12B-IT\\ $s = 16.7\%$\end{tabular}} & \multirow{2}{*}{\rotatebox[origin=c]{90}{\scriptsize Baseline}} & \textcolor{refgray}{\textit{Dense}} & 0.53 & 0.84 & 0.51 & 0.86 & 0.44 & 0.95 & 0.24 & 0.61 & 0.07 & 0.71 & 0.00 & 0.40 & 0.83 & 0.73 & 0.74 & 0.63 & 0.35 & 0.74 & 0.60 \\
 &  & BI & 0.25 & 0.45 & 0.28 & 0.54 & 0.00 & 0.01 & 0.02 & 0.37 & 0.00 & 0.22 & 0.00 & 0.19 & 0.63 & 0.53 & 0.49 & 0.00 & 0.23 & 0.50 & 0.31 \\
\addlinespace[1.5pt]
\cmidrule(lr){2-22}
\addlinespace[1.5pt]
 & \cellcolor{ourstint}{} & \cellcolor{ourstint}{J\textsubscript{Mag}} & \cellcolor{ourstint}{0.19} & \cellcolor{ourstint}{0.34} & \cellcolor{ourstint}{0.29} & \cellcolor{ourstint}{0.56} & \cellcolor{ourstint}{0.00} & \cellcolor{ourstint}{0.01} & \cellcolor{ourstint}{0.00} & \cellcolor{ourstint}{0.25} & \cellcolor{ourstint}{0.00} & \cellcolor{ourstint}{0.24} & \cellcolor{ourstint}{0.00} & \cellcolor{ourstint}{0.14} & \cellcolor{ourstint}{0.54} & \cellcolor{ourstint}{0.52} & \cellcolor{ourstint}{0.47} & \cellcolor{ourstint}{0.00} & \cellcolor{ourstint}{0.24} & \cellcolor{ourstint}{0.53} & \cellcolor{ourstint}{0.29} \\
 & \cellcolor{ourstint}{} & \cellcolor{ourstint}{J\textsubscript{Logit}} & \cellcolor{ourstint}{0.20} & \cellcolor{ourstint}{0.50} & \cellcolor{ourstint}{0.29} & \cellcolor{ourstint}{0.53} & \cellcolor{ourstint}{0.00} & \cellcolor{ourstint}{0.01} & \cellcolor{ourstint}{0.01} & \cellcolor{ourstint}{0.27} & \cellcolor{ourstint}{0.00} & \cellcolor{ourstint}{0.22} & \cellcolor{ourstint}{0.00} & \cellcolor{ourstint}{0.12} & \cellcolor{ourstint}{0.60} & \cellcolor{ourstint}{0.58} & \cellcolor{ourstint}{0.48} & \cellcolor{ourstint}{0.00} & \cellcolor{ourstint}{0.24} & \cellcolor{ourstint}{0.50} & \cellcolor{ourstint}{0.30} \\
 & \cellcolor{ourstint}{} & \cellcolor{ourstint}{J-BI\textsubscript{Mag}} & \cellcolor{ourstint}{\textbf{0.47}} & \cellcolor{ourstint}{\textbf{0.74}} & \cellcolor{ourstint}{0.44} & \cellcolor{ourstint}{\textbf{0.87}} & \cellcolor{ourstint}{\textbf{0.29}} & \cellcolor{ourstint}{\textbf{0.30}} & \cellcolor{ourstint}{\textbf{0.17}} & \cellcolor{ourstint}{\textbf{0.44}} & \cellcolor{ourstint}{\textbf{0.05}} & \cellcolor{ourstint}{\textbf{0.66}} & \cellcolor{ourstint}{0.00} & \cellcolor{ourstint}{0.32} & \cellcolor{ourstint}{0.72} & \cellcolor{ourstint}{0.70} & \cellcolor{ourstint}{0.71} & \cellcolor{ourstint}{0.22} & \cellcolor{ourstint}{\textbf{0.28}} & \cellcolor{ourstint}{\textbf{0.69}} & \cellcolor{ourstint}{\textbf{0.51}} \\
 & \cellcolor{ourstint}{\multirow{-4}{*}{\rotatebox[origin=c]{90}{\scriptsize \bfseries Ours}}} & \cellcolor{ourstint}{J-BI\textsubscript{Logit}} & \cellcolor{ourstint}{0.44} & \cellcolor{ourstint}{0.72} & \cellcolor{ourstint}{\textbf{0.51}} & \cellcolor{ourstint}{0.84} & \cellcolor{ourstint}{0.27} & \cellcolor{ourstint}{0.12} & \cellcolor{ourstint}{0.16} & \cellcolor{ourstint}{\textbf{0.44}} & \cellcolor{ourstint}{0.04} & \cellcolor{ourstint}{0.56} & \cellcolor{ourstint}{0.00} & \cellcolor{ourstint}{\textbf{0.33}} & \cellcolor{ourstint}{\textbf{0.73}} & \cellcolor{ourstint}{\textbf{0.73}} & \cellcolor{ourstint}{\textbf{0.72}} & \cellcolor{ourstint}{\textbf{0.32}} & \cellcolor{ourstint}{\textbf{0.28}} & \cellcolor{ourstint}{0.68} & \cellcolor{ourstint}{\textbf{0.51}} \\
\addlinespace[1.5pt]
\bottomrule
\end{tabular}}
\caption{\textbf{Per-task accuracy with $k=8$ blocks removed.} $s$ is the depth sparsity $k/n_\mathrm{layers}$, constant within a model group ($s=0$ for \textit{Dense}). The \emph{Baseline} rows are the reference points: \textit{Dense} is the unpruned model and BI is the published block-influence pruner from ShortGPT. \textbf{Ours} rows are the Jacobian-informed metrics introduced in this work. \textbf{Bold} marks the best pruning metric for each task and model at the precision shown. In the case of a tie all metrics are marked. \emph{Average} is the macro-average over the task units admitted for that model.   Datasets were admitted into a model's average only if the dense model had 5pp or higher accuracy to reduce noise effects.  Thus the Average here is not a strict average of all row items. Filtered datasets are still shown here for completeness as they may be included at other $k$.}

\label{tab:acc_k8}
\end{table}

\section{Block Removal Allocation}\label{block_remove}

\begin{figure}
    \centering
    \includegraphics[width=0.9\linewidth]{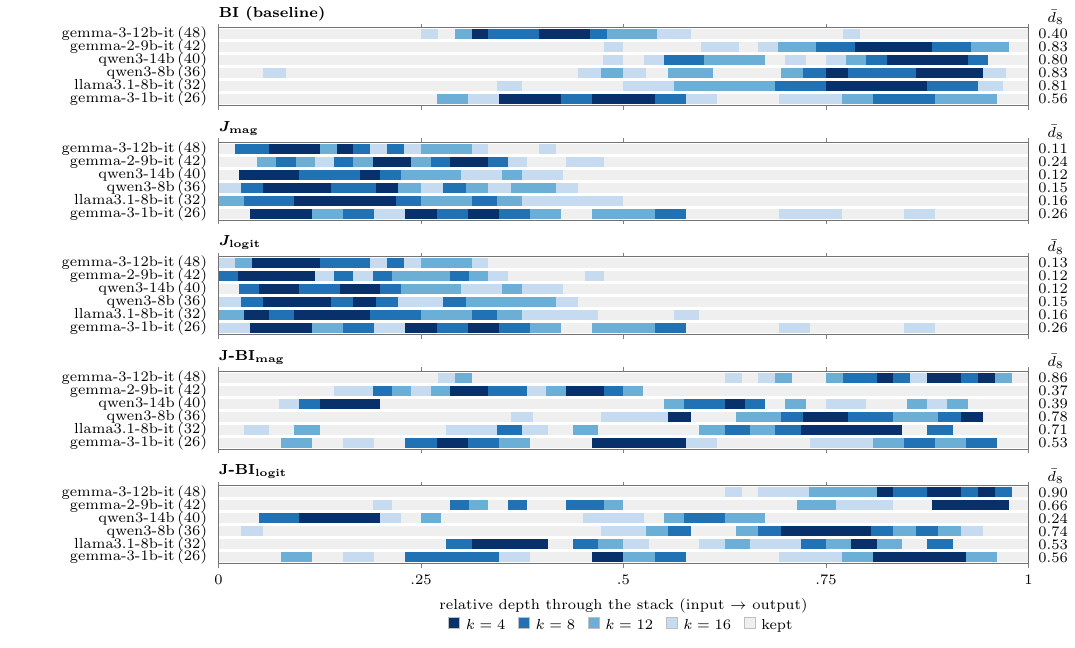}
    \caption{Layer removal allocation by scoring method across a range of layers removed. Evaluated on Gemma-3-12b and 1b, Qwen3-14b and 8b, and Llama3.1-8b and Gemma-2-9b models, using Block Influence calibration set}
    \label{fig:appendix-block_removal_alloc}
\end{figure}

\begin{figure}
    \centering
    \includegraphics[width=0.9\linewidth]{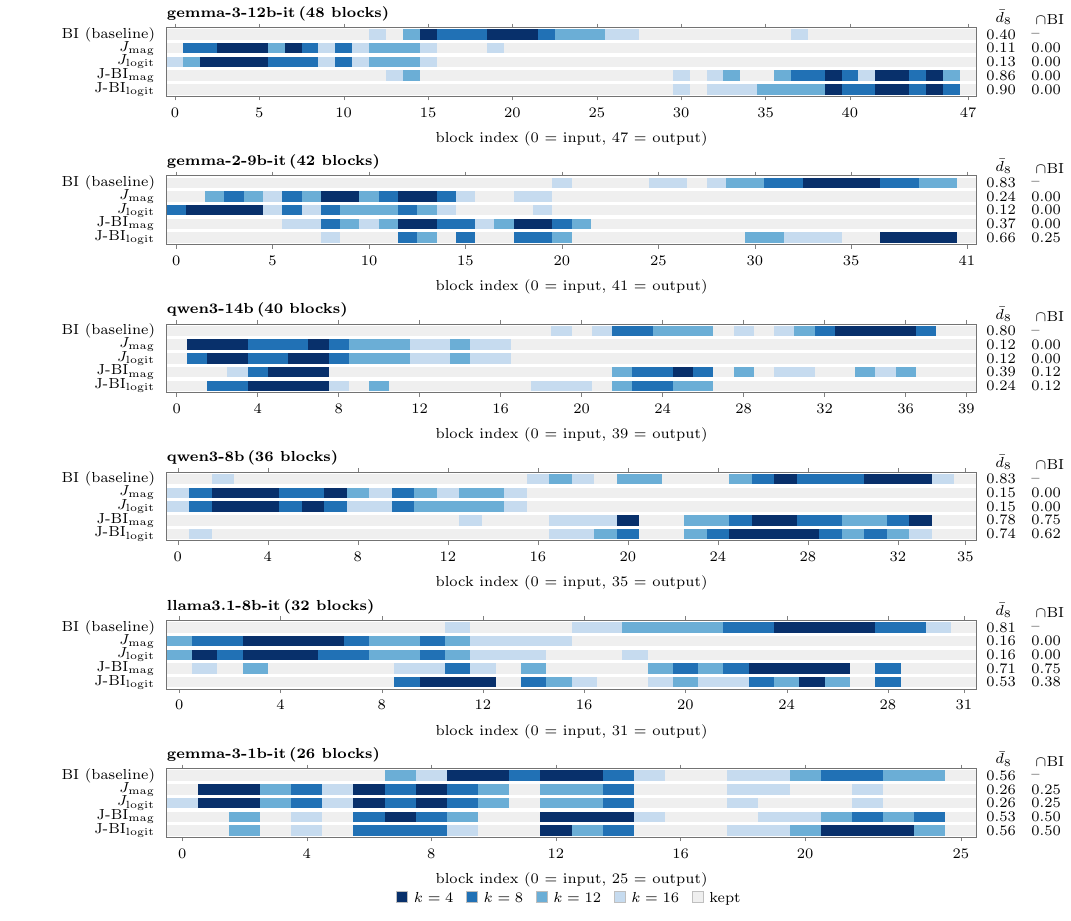}
    \caption{Layer removal allocation by model across a range of layers removed. Evaluated on Gemma-3-12b and 1b, Qwen3-14b and 8b, and Llama3.1-8b and Gemma-2-9b models, using Block Influence calibration set}
    \label{fig:appendix-block_removal_model_alloc}
\end{figure}
\paragraph{Where each score prunes.}
Here we compare where each score in Section~\ref{sec:method} removes blocks, relative to BI. Every score is computed on the same calibration set, and the number of removed blocks is varied from 4 to 16.

As Figure~\ref{fig:appendix-block_removal_alloc} shows, BI tends to prune mid-to-deep blocks and generally avoids early ones. At $k=8$ its mean pruning depth is about 80\% on the non-Gemma-3 models, but only about 50\% on Gemma-3.

$\JMag$ and $\JLogit$ both prune much shallower blocks across models. At $k=8$ their mean pruning depth over all models is 17\% and 16\% respectively. Removing early blocks is costly (Figure~\ref{fig:single-block}), which is consistent with these scores collapsing first.

$\JBIMag$ and $\JBILogit$ prune across a wider range of depths than the lens-only scores. At $k=8$, $\JBIMag$ removes blocks from 37\% to 86\% depth and $\JBILogit$ from 24\% to 90\%. Their overlap with BI is also much higher, especially on the non-Gemma models (Figure~\ref{fig:appendix-block_removal_model_trio_alloc}). On Qwen3-8B and Llama-3.1-8B, $\JBIMag$ shares 75\% of its removed blocks with BI at $k=8$; $\JBILogit$ shares 62\% on Qwen3 and 38\% on Llama-3.1.

\begin{figure}
    \centering
    \includegraphics[width=0.8\linewidth]{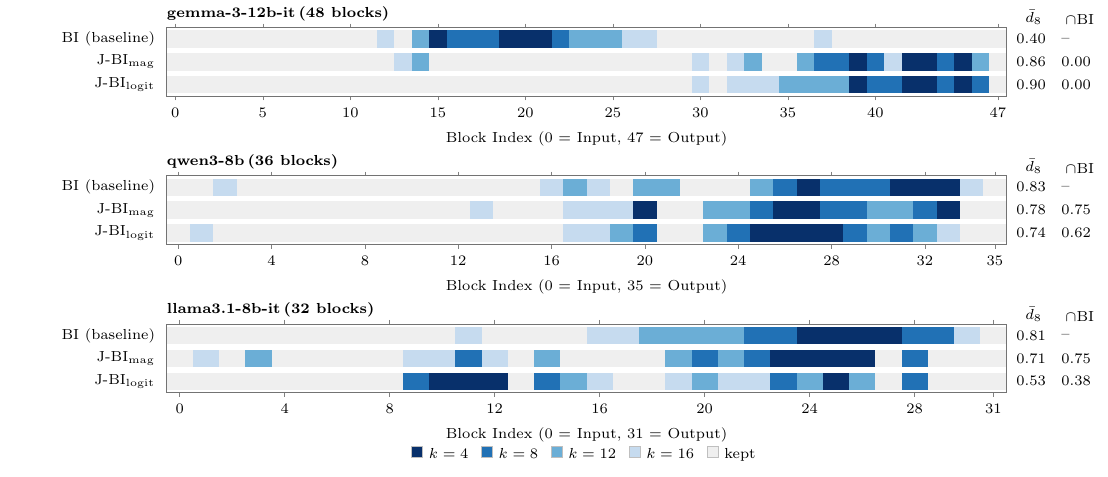}
    \caption{Block removal allocation by model across a range of removed blocks, for Gemma-3-12B, Qwen3-8B and Llama-3.1-8B, using the Block Influence calibration set.}
    \label{fig:appendix-block_removal_model_trio_alloc}
\end{figure}

\section{Why J-BI removes different blocks: a lens readout}
\label{jbi_arch_agnostic}

Measuring BI at the output was expected to help accuracy. It does not by itself explain why the J-BI scores also give much better perplexity. Here we use the J-Lens as it was designed, as an interpretability tool, to propose an explanation based on which blocks each score removes.

\begin{figure}
    \centering
    \includegraphics[width=1.0\linewidth]{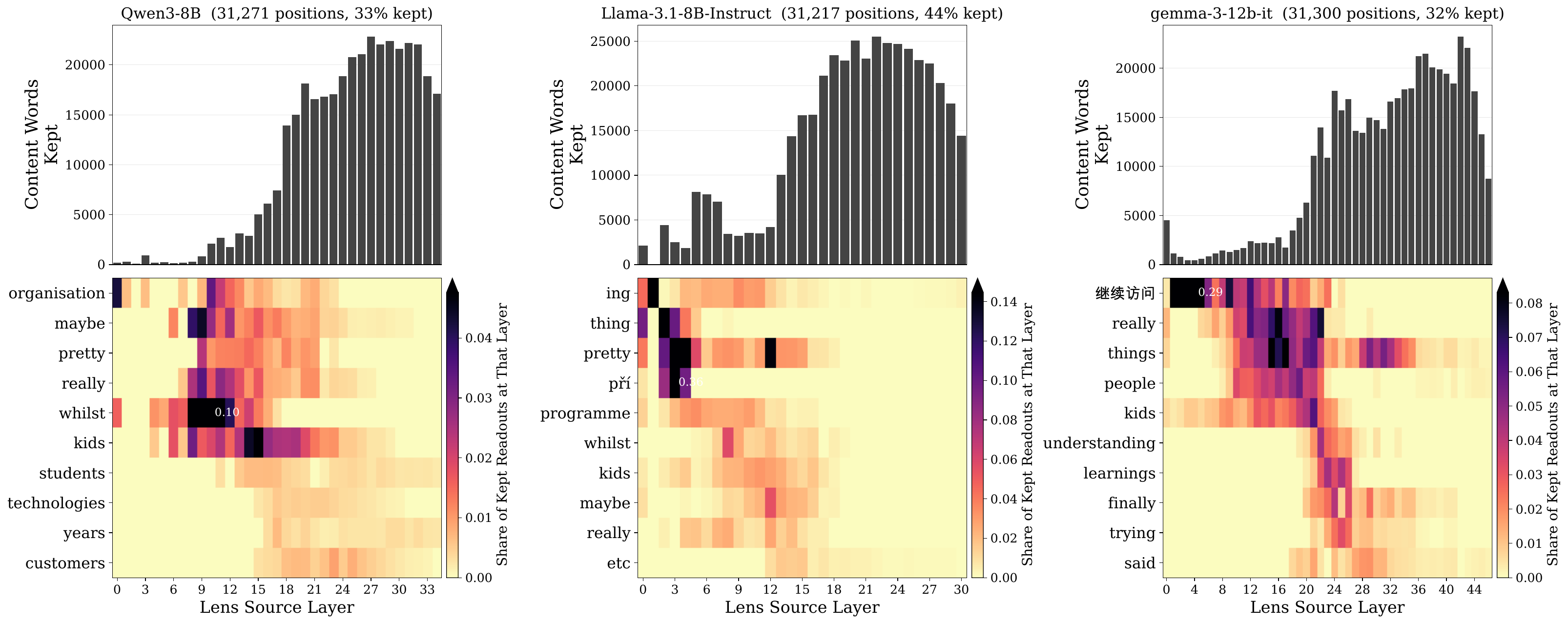}
    \caption{(\emph{Top}) Where the model's next-token predictions first take shape. Frequency of readable J-Lens token readouts at each layer, after filtering out stop-words, junk tokens and punctuation, for Qwen3-8B, Llama-3.1-8B and Gemma-3-12B on C4. (\emph{Bottom}) Distribution over layers of the most frequent lens tokens; for example ``really'', ``understanding'' and ``learnings'' all appear frequently in the middle layers of Gemma.}
    \label{fig:results-token_freq_lens}
\end{figure}

The lens reads out, at every layer, the next token the model is heading toward. Figure~\ref{fig:results-token_freq_lens} shows that each model first surfaces readable lens tokens in a different part of the network. Qwen3-8B does so in the early-middle layers (9 to 18). Llama-3.1-8B does so in the early layers (4 to 7) and again from layer 12 onward. Gemma-3-12B does so in a middle group of blocks (20 to 28). If these are the blocks where predictions take shape, a score should avoid removing them.

Removing blocks one at a time (Figure~\ref{fig:single-block}) confirms this. The blocks the lens marks as important are the blocks that removal hurts most, and they are the blocks the J-BI scores avoid and BI may prune. On Qwen3-8B and Llama-3.1-8B, BI and J-BI largely agree: all scores remove deep blocks, where the lens sees few distinct tokens. On Gemma-3-12B, BI prunes the important middle blocks, while J-BI avoids them in favour of the final blocks.

One reason for the difference may be Gemma-3-12B's hybrid attention. Every sixth layer attends to the global context; the layers in between use sliding-window attention. BI sees that the middle blocks' inputs and outputs are cosine-similar, but it cannot see that the rest of the model is sensitive to what those blocks write. J-BI measures the same angle after transport to the output, where that sensitivity is visible. We offer this as a hypothesis rather than an architecture-wide claim; the evidence here is a single model. If it holds, J-BI should outperform BI on a wider range of families for the same reason.

\end{document}